# A Comparison of Machine Learning Algorithms for the Surveillance of Autism Spectrum Disorder

Scott Lee, Matthew Maenner, Chad Heilig
Centers for Disease Control and Prevention, Atlanta, GA

# Abstract
## Objective
The Centers for Disease Control and Prevention (CDC) coordinates a labor-intensive process to measure the prevalence of autism spectrum disorder (ASD) among children in the United States. Random forests methods have shown promise in speeding up this process, but they lag behind human classification accuracy by about 5%. We explore whether more recently available document classification algorithms can close this gap.

## Materials and methods
We applied 8 supervised learning algorithms to predict whether children meet the case definition for ASD based solely on the words in their evaluations. We compared the algorithms' performance across 10 random train-test splits of the data, using classification accuracy, $F_1$ score, and number of positive calls to evaluate their potential use for surveillance.

## Results
Across the 10 train-test cycles, the random forest and support vector machine with Naive Bayes features (NB-SVM) each achieved slightly more than 87% mean accuracy. The NB-SVM produced significantly more false negatives than false positives ($P = 0.027$), but the random forest did not, making its prevalence estimates very close to the true prevalence in the data. The best-performing neural network performed similarly to the random forest on both measures.

## Discussion
The random forest performed as well as more recently available models like the NB-SVM and the neural network, and it also produced good prevalence estimates. NB-SVM may not be a good candidate for use in a fully-automated surveillance workflow due to increased false negatives. More sophisticated algorithms, like hierarchical convolutional neural networks, may not be feasible to train due to characteristics of the data. Current algorithms might perform better if the data are abstracted and processed differently and if they take into account information about the children in addition to their evaluations.

## Conclusion

Deep learning models performed similarly to traditional machine learning methods at predicting the clinician-assigned case status for CDC's autism surveillance system. While deep learning methods had limited benefit in this task, they may have applications in other surveillance systems.

## Introduction

The Centers for Disease Control and Prevention (CDC) coordinates a labor-intensive process to measure the prevalence of autism spectrum disorder (ASD) among children in the United States. Maenner et al.[1] developed a promising machine learning approach that could assist with portions of this process. In this paper, we expand on this initial work by evaluating a wider variety of machine learning models.

ASD (here used interchangeably with "autism") refers to a group of heterogeneous neurodevelopmental conditions characterized by impairments in social interaction and the presence of repetitive behaviors or restricted interests. ASD is diagnosed through the observation of behavior consistent with the criteria described in the *Diagnostic and Statistical Manual of Mental Disorders*.[2] Since 2000, CDC has monitored the prevalence of ASD among US children in selected communities through the Autism and Developmental Disabilities Monitoring (ADDM) Network using a process by which trained clinicians review children's medical and educational evaluations to identify behaviors consistent with the DSM criteria for ASD. The surveillance case definition, which serves a different purpose than a medical diagnosis, allows the ADDM Network to identify children who have descriptions of the requisite behavioral features documented in their records, but do not necessarily have an ASD diagnosis. Every two years, the ADDM Network has used this method to estimate the prevalence of ASD in 8-year-old children, ranging from 1 in 150 children in 2000 to 1 in 68 children in 2012. The ADDM network has yielded crucial insights into the epidemiology of ASD in terms of understanding prevalence, disparities in diagnosis, and the contribution of risk factors to the changes in prevalence over time.

Because expert clinicians must manually review each child's evaluations to determine whether they meet the surveillance case definition, the system is both labor-intensive and time-consuming. To explore ways of making the review process more efficient, Maenner et al.[1] developed a machine learning algorithm for automatically determining whether children meet the ADDM surveillance case definition for ASD based solely on the text contained in their written evaluations. By training a random forest[3] on written evaluations collected in 2008, they were able to predict classifications for evaluations collected in 2010 with good diagnostic accuracy, achieving an $F_1$ score of 86.6% and an accuracy score of 86.5%, compared to interrater agreement among the expert reviewers of 90.2%[4]; accuracy and $F_1$ are further explained in the methods section. The algorithm could also be used by ADDM clinicians to screen children during the manual review process and to focus their efforts on cases that are harder to classify, where good judgment and clinical experience are critical for classification.

We conceived our study to expand upon Maenner et al.'s random forest analysis in order to achieve three primary goals. First, we sought to determine whether we can achieve higher accuracy on the case classification task using more recently available analytical methods, including those falling under the broad umbrella of deep learning. Second, we wanted to assess the variability in performance of these methods, as Maenner et al. considered only 1 train-test split. Finally, we aimed to compare differences in the prevalence estimates produced by these methods, which has direct bearing on their suitability for surveillance. We discuss ways in which these models may be used effectively to enhance autism surveillance.

# Methods
## ADDM Overview

Maenner et al.[1] provide a detailed overview of the ADDM Network, including the labor-intensive review process used to determine whether children meet the surveillance case definition for ASD, and an explanation of how machine learning algorithms may be used to assist clinicians in conducting the manual reviews. In brief, each site in the network requests to review medical records for children evaluated for having developmental disabilities and educational records for children served in a special education program. These records are screened by ADDM Network staff. If a record contains a possible indication of autism (including a diagnosis, specific behaviors, and if an autism test was given), the text from the child's developmental evaluations are extracted into the surveillance database. Evaluations from multiple sources are combined into a single, de-identified record and reviewed by trained clinicians to determine if they meet the ADDM Network ASD case definition. Because the focus of this study is on the comparison of methods for document classification rather than their implementation in the ADDM surveillance workflow, we refer readers to other sources[5,6] for more information on the structure and goals of the network.

## Corpus and Data Structure

Our dataset consists of the abstracted evaluations and corresponding surveillance case classifications for all children evaluated in years 2006[7], 2008[4], and 2010[8] at the Georgia ADDM site. During these 3 waves, 3,379 children were reviewed at the site, among which 1,829 (48.9%) met the ADDM surveillance case definition for ASD. Our analytic dataset is a corpus $D$ of 3,739 documents, with a vocabulary size $V$ of 59,660 and a total word count $W$ of 7,845,838. The documents range in length from only a couple of words to the tens of thousands (Table 1).

|  | **Min** | **1Q** | **Med** | **3Q** | **Max** |
|---|---|---|---|---|---|
| *Total* | 2 | 813 | 1,528 | 2,737 | 20,801 |
| *Unique* | 2 | 344 | 527 | 773 | 2407 |

Table 1. Minimum, first quartile, median, third quartile, and maximum word counts per child. The first row shows statistics for total (i.e., non-unique) words, while the second shows those for unique words. We represented each child's record as the unordered collection of his or her abstracted evaluations, which we treated as a single block of text (i.e., a document). We preprocessed the text by lowercasing all strings, removing stop words and special characters, and converting all words to their dictionary forms, or lemmas.

For our baseline classifiers, we represented each child's collection of abstracted evaluations as a single document in a bag-of-words (BoW) model. Under this model, each document $d$ is represented as a row vector of word counts, where each entry in the row corresponds to the number of times a particular word *w* appears in the document. The entire corpus is represented as a *d* x *V* document-term matrix, where each row is the BoW vector for a particular child's combined abstracted evaluations. To make our classifiers more effective, we counted both single words, or unigrams (n=59,660), and pairs of adjacent words, or bigrams (n=830,803); this yielded a total of 890,463 features in our data representation. BoW classification models are computationally efficient and readily applied using widely available, open-source software. In addition, some classifiers applied to BoW data can yield metrics interpretable as feature importances, which can give investigators useful clues as to how the model learns to discriminate cases from non-cases.

## Description of Classifiers

We compared several baseline classifiers to the random forest algorithm published by Maenner et al.[1]: latent Dirichlet allocation (LDA)[9,10]; latent semantic analysis (LSA)[11]; multinomial naive Bayes (MNB)[12]; support vector machine (SVM) with a linear kernel[13]; interpolated naive Bayes-SVM (NB-SVM)[14,15]; and two neural networks adapted from the fastText architecture[16].

Latent Dirichlet allocation (LDA)[9] is an unsupervised algorithm typically used for topic modeling rather than document classification. LDA models documents as mixtures of topics, which themselves are modeled as mixtures of words, allowing it to model complex and often subtle information in large collections of text. LDA has been adapted with some success for supervised learning problems.[10] We consider LDA as a way to generate dense vector representations of the evaluations, which can then be used as input for training a supervised algorithm. Latent semantic analysis[11] is a dimensionality reduction technique that generates dense representations of the evaluations by applying a singular value decomposition to the document-term matrix. For both LDA and LSA, we used a linear SVM [13] to perform the case classification task.

Multinomial Naive Bayes (NB)[12] is a supervised learning algorithm that uses Bayes' rule to calculate the probability that a document belongs to a certain class based on the words (also known as features) that it contains, under the assumption that the features are statistically independent conditional on class membership. It is often used as a baseline model for text classification. Multinomial NB produces the most likely features for each class of documents, which can yield keywords associated with evaluations meeting the surveillance case definition for ASD. The model can also generate predicted class probabilities to use for classification.

While multinomial NB models are interpretable and quick to train, they have some formal shortcomings,[12] like the conditional independence assumption mentioned above, and they are often outperformed by discriminative models like the support vector machine (SVM). For this reason, we also included 2 versions of the SVM using the document-term matrix as input: a simple linear-kernel SVM, and an interpolated Naive Bayes-SVM (NB-SVM).[14] The SVM constructs a maximum-margin decision boundary between document classes based on the original document-term matrix. The NB-SVM constructs a decision boundary using NB features, which makes it competitive with state-of-the-art models for sentiment analysis.[14] The model tends to work best when nonzero word counts are converted to 1, or binarized. This change makes the weights in the trained model heuristically (but not strictly) interpretable as a kind of feature importance.

Our final models are both neural networks and are simplified versions of the fastText architecture.[16] Like the NB-SVM, they take a binarized document-term matrix as their input, and like a traditional logistic regression model, they output class probabilities via the softmax function that can be used for document classification. Unlike the other models in our experiments, the networks feature an embedding layer between the input and softmax layers, allowing them to learn dense vector representations of words rather than documents. In the original fastText architecture, these vectors are averaged to generate document representations; we reuse this method for our first model ($NN_{avg}$), and we replace the averaging layer with a summation layer for our second ($NN_{sum}$).

A supplement provides additional technical details about the model architectures, hyperparameters, and implementation.

## Hyperparameter Optimization

Before performing our experiment, we randomly split the full dataset into a training set and a validation set, which we then used to select hyperparameter values for each model. We used a variety of methods for tuning, including grid search (LSA and LDA), a combination of non-recursive and recursive feature elimination (the random forest), and a Bayesian optimization procedure based on Gaussian processes (all other models). We provide detailed descriptions of the optimization procedure for each model in the Supplement.

## Experimental Setup

Maenner et al.[1] mimicked real-world conditions by training their model on data gathered from the Georgia ADDM site in 2008 and then testing it on data collected in 2010. Because we were interested in assessing both the performance and the variability in performance of our models, we formulated our experiment as a series of 10 train-test cycles, where the training data are selected randomly from the entire dataset rather than by year. For each of these cycles, we randomly split the entire dataset into 57% training, 13% validation, and 30% test sets. We then fit each model to the training data, and measured its performance on the test data using common measures of binary classification accuracy, including raw accuracy (the proportion correctly classified) and $F_1$ score (the harmonic mean of sensitivity and positive predictive value). Because public health surveillance relies on accurate prevalence estimates, we also measure the difference between each model's number of positive calls and the true number of cases in the test set. We used a fixed list of 10 seeds for the random number generator to ensure that the models were fit and tested on exactly the same data splits.

To compare the performance of the models across the 10 train-test splits, we used the Wilcoxon signed-rank test for paired data. We focused on two metrics: mean classification accuracy (individual-level prediction), and mean difference in prevalence from the true prevalence in the test data (population-level prediction). For each metric, we selected the model with the best score as the referent, and we conducted multiple pairwise comparisons between it and the remaining models to determine significant differences. We use the Benjamini-Yekutieli[17] procedure to control for the false discovery rate, and we report the adjusted p-values in our results, considering values of less than 0.05 to denote statistical significance.

## Technical Notes

The LDA, LSA, multinomial NB, SVM, and random forest models were implemented in Python using scikit-learn v0.19,[18] which was also used to preprocess the text and generate the document-term matrices. The NB-SVM was implemented in NumPy,[19] with the SVM component imported from scikit-learn, and the neural networks were implemented in Keras with the TensorFlow backend.[20] Bayesian hyperparameter optimization was implemented using GPyOpt. Finally, statistical analysis was conducted in R 3.5.1 [21].

This analysis was submitted for human subjects review and deemed to be non-research (public health surveillance) according to CDC policy.

## Data Availability

The primary data in this analysis are medical and educational evaluations collected for public health surveillance under an assurance of confidentiality pursuant to the Public Health Service Act, §308(d). Due to the sensitive nature of these documents, we will make these data available (upon request) in the form of the final term-document matrices used to train and test the models' performance rather than the raw text of the evaluations; these matrices will not include an enumeration of the *n*-grams associated with the features, and so they will be purely numeric. CDC's National Center on Birth Defects and Developmental Disabilities requires a signed data use agreement by anyone requesting data from the Metropolitan Atlanta Developmental Disabilities Surveillance Program (MADDSP) to ensure that: 1) the data are analyzed for the specific purpose of the proposal submitted, and 2) the investigator will not try to identify any child or present stratified analyses leading to a sample <5 children. These two points are what result in the dataset being considered a restricted public use dataset. All requests for MADDSP public use datasets should be submitted to ncbddddata@cdc.gov.

## Code Availability

The code for our models, optimization procedures, and experiments is available on GitHub at https://github.com/scotthlee/autism_classification/.

## Results

We present the mean binary classification metrics for each of our models across the 10 train-test splits in Table 2.

The NB-SVM achieved the highest mean accuracy (87.62%) across the 10 train-test cycles, followed by the random forest (87.07%), the averaging neural network (86.30%), and the summing neural network (85.08%). The mean $F_1$ scores were also very close, with the top 2 models, the NB-SVM (87.07%) and the random forest (86.81), being separated by only a quarter of a percentage point; these 2 models also achieved the highest scores for sensitivity, specificity, PPV, and NPV. Although five models achieved accuracy of over 85%, the random forest was the only model that was not significantly different from the NB-SVM in terms of accuracy ($p$=0.090).

| Model | Sens | Spec | PPV | NPV | F$_1$ | Acc | Acc $p$ (adj) |
|---|---|---|---|---|---|---|---|
| LDA | 44.19 | 72.39 | 60.56 | 57.51 | 51.08 | 58.59 | 0.018 |
| MNB | 82.26 | 72.60 | 74.22 | 81.04 | 78.02 | 77.33 | 0.018 |
| SVM | 83.46 | 84.47 | 83.75 | 84.22 | 83.59 | 83.98 | 0.018 |
| LSA | 81.46 | 88.50 | 87.20 | 83.30 | 84.21 | 85.05 | 0.018 |
| NN$_{sum}$ | 85.50 | 84.68 | 84.43 | 86.03 | 84.87 | 85.08 | 0.018 |
| NN$_{avg}$ | 86.25 | 86.35 | 85.93 | 86.88 | 86.02 | 86.30 | 0.018 |
| RF | **87.01** | 87.12 | 86.64 | **87.53** | 86.81 | 87.07 | 0.090 |
| NB-SVM | 85.23 | **89.91** | **89.03** | 86.42 | **87.07** | 87.62 | * |

Table 2. Mean performance for our 8 models across the 10 train-test splits. Metrics include sensitivity (Sens), specificity (Spec), positive predictive value (PPV), negative predictive value (NPV), F$_1$, and accuracy (Acc); the best scores for each metric are shown in bold. The final column presents multiplicity-adjusted p-values from a Wilcoxon signed-rank test comparing each model to the NB-SVM, with statistical significance set at $p < 0.05$.

Although our classifiers yielded similar accuracy, they differed in their proportions of positive calls, as well as in the distribution of their incorrect predictions between positive and negative calls. The random forest and the two neural networks produced about as many false positives (FPs) as false negatives (FNs), with mean proportions positive of 48.41% and 48.65% respectively (Table 3).

The NB-SVM and LSA models, however, leaned more heavily on FNs than FPs, with mean differences of -23 and -36 respectively in the number of positives from the test set. On the other hand, MNB produced many more FPs than FNs, resulting in a higher mean prevalence estimate than the true proportion.

| Model | FP | FN | $n$ pos | Diff pos | p | Pair. p |
|---|---|---|---|---|---|---|
| LDA | 158 | 306 | 401 | -148 | 0.006 | 0.027 |
| MNB | 157 | 97 | 609 | 60 | 0.006 | 0.027 |
| SVM | 89 | 91 | 547 | -2 | 0.721 | * |
| LSA | 66 | 102 | 513 | -36 | 0.006 | 0.027 |
| NN$_{sum}$ | 88 | 80 | 557 | 8 | 0.610 | 1.000 |
| NN$_{avg}$ | 78 | 76 | 552 | 3 | 0.790 | 1.000 |
| RF | 74 | 71 | 552 | 3 | 0.139 | 0.504 |
| NB-SVM | 58 | 81 | 526 | -23 | 0.006 | 0.027 |

Table 3. Mean prevalence-related metrics for our 8 models across the 10 train-test splits. Metrics included are false positives (FP); false negatives (FN); number of positive calls ($n$ pos); difference between number positive calls and the number of actual positives in the test set (Diff pos); p-values for a one-sample Wilcoxon signed rank test for Diff pos being centered on 0; and p-values from two-sample Wilcoxon signed rank tests comparing each model's Diff pos to that of the SVM.

# Discussion

Our baseline models are strong to enough enhance the current surveillance workflow: Their accuracy is within 5% of human levels on the same task, they are computationally feasible, and they are heuristically interpretable. As we discuss here, more sophisticated models alone cannot be expected to improve performance without enriching the representation of the data, e.g., by way of feature engineering, richer representation of text than unigram and bigram bag-of-words, or including other information from the children's records in the model.

Perhaps our most important result is that the random forest was statistically indistinguishable from the NB-SVM in its individual-level performance (i.e. its classification accuracy) and from the SVM in its population-level performance (i.e. its prevalence estimate). Especially given the interpretability of its feature importances, these two results suggest that the random forest stands out as a good candidate for surveillance applications among the models that we evaluated.

For surveillance purposes, accuracy or $F_1$ scores may have less practical importance than the number of positive calls, which public health practitioners often use to generate model-based prevalence estimates. In a fully-automated workflow, then, the random forest or neural network may be more appropriate for conducting surveillance, since they produce more accurate prevalence estimates without sacrificing much in the way of individual-level predictive quality. As a bonus, these two methods also naturally produce predicted class probabilities, which could be used to support the current surveillance workflow by helping clinicians focus on evaluations that may be particularly hard to classify. In a partially-automated setting, however, the NB-SVM may still be useful as a support tool for clinicians conducting a manual review of the written evaluations, especially if cross-validation (e.g., by way of Platt scaling[22]) is used to obtain non-thresholded probability estimates that are well-calibrated to the true distribution of class labels.

Another important result is that none of the models was able to match the levels of interrater agreement seen in the ADDM network's ongoing quality reviews[4], although both the random forest and the NB-SVM achieved over 89% on several train-test cycles. In broad terms, this result suggests that the clinicians reviewing the evaluations rely on more than just the text they contain to determine whether a child meets the surveillance case definition for ASD. In practice, the ADDM clinicians have access to more than just the written evaluations when making their case classifications. Because interrater agreement among the clinicians hovers around 91%, we would likely need to add extra features to the analytic data to lower the error rate, regardless of which document-level classifier is used. Maenner et al.[1] made note of this in their original analysis, noting three possible refinements beyond document-level models to improve classification: (1) accounting for characteristics of each child's set of evaluations (such as total number and mix of school or health sources), (2) adding phrase-level information to the document-level classifiers to approximate the symptom-based scoring rubric used by the clinicians, or (3) using additional characteristics of the children themselves, such as sex or IQ. Since our purpose in this analysis was to compare alternative document-level classifiers, we did not assess the potential incremental improvements from using other features. Based on our results, we conclude that using additional features would be the logical way to further reduce the gap between a machine algorithm and clinical interrater agreement.

To address the question of whether more sophisticated text-classification models could achieve higher levels of accuracy on this particular task, we refer back to the child-level descriptive statistics for the corpus (Table 1), which demonstrate two important characteristics of the ADDM dataset: variability in length of the abstracted evaluations, and variability of their vocabularies. The BoW model is able to accommodate this variability in a straightforward way, through the construction of the document-term matrix and its variants, but it may pose a challenge for other classification models. Recurrent neural networks (RNNs) can have a hard time learning long sequences due to the vanishing/exploding gradient problem.[23] Long short-term memory networks[24] and gated recurrent units[25] generally solve this problem by altering the standard RNN cell so that it forgets information that is unimportant for prediction, would be unlikely to classify the longest documents in our dataset without substantial modification. Convolutional neural networks (CNNs) have also been used for text classification,[26,27] but they do not appear to work well for longer chunks of text without substantial modification. Denil et al.[28] used a hierarchical CNN to generate document representations from lower-level information in the text, like words and sentences. These and other sophisticated models, like a recurrent CNN,[29] a gated recurrent network,[30] and paragraph vectors[31] may achieve higher levels of classification accuracy on this particular task. They may not be worth the effort to implement, however, given our results. Our baseline classifiers have simpler architectures, are far less computationally intensive, and produce relatively unbiased prevalence estimates, all without sacrificing much in the way of individual-level prediction.

On a practical note, public health practitioners should carefully consider what they hope to achieve by applying machine learning to surveillance, and they should choose models that will help them achieve these specific goals. In low-resource settings where continuing expert review is infeasible and the model alone will be used to generate prevalence estimates, diagnostic accuracy may be less important than similarity between the proportion of positive calls produced by the model and the class labels in the actual data. Statistical methods for paired proportions, like McNemar's test or Newcombe's[32] method for estimating the corresponding confidence intervals, can be used in these contexts to judge the quality of predictions.[33,34] In higher-resource settings where expert review is a component of the surveillance system, as in the ADDM network, probabilistic calibration through measures like the Brier score or cross-entropy loss becomes more important, since reviewers can use the model-based probability estimates to focus their efforts on cases that are hard to classify. Sensitivity, specificity, and other measures of binary diagnostic accuracy are still useful, especially when models are used for patient-level screening or diagnosis, but when models are used for population-level surveillance, the other measures bear careful consideration.

# Conclusion

Although more sophisticated models do not appear to be necessary for improving the autism surveillance workflow, these and other deep models could be useful in the general sense for other public health applications; CDC, for example, maintains several large databases containing unstructured text for which these methods might improve the efficiency of surveillance systems.

# Disclaimer

# Acknowledgements

This project was supported by contributions from several individuals and groups, but most especially the Georgia site of the ADDM Network and the ADDM data team at CDC, who procured the data.

# References

1  Maenner MJ, Yeargin-Allsopp M, Braun KV, et al. Development of a machine learning algorithm for the surveillance of autism spectrum disorder. PLOS ONE. 2016 Dec 21;11(12):e0168224.
2  American Psychiatric Association. (2000). *Diagnostic and Statistical Manual of Mental Disorders: DSM-IV-TR*. Washington, DC: American Psychiatric Association.
3  Breiman L. Random forests. Machine learning. 2001 Oct 1;45(1):5-32.
4  Autism and Developmental Disabilities Monitoring Network Surveillance Year 2008 Principal Investigators. Prevalence of autism spectrum disorders—Autism and Developmental Disabilities Monitoring Network, 14 sites, United States, 2008. MMWR Surveill Summ 2012;61(No. SS-3):1–19.
5  Rice CE, Baio J, Van Naarden et al. A public health collaboration for the surveillance of autism spectrum disorders. Paediatric and Perinatal Epidemiology. 2007 Mar 1;21(2):179-90.
6  Christensen DL, Baio J, Braun KV, et al. Prevalence and Characteristics of Autism Spectrum Disorder Among Children Aged 8 Years — Autism and Developmental Disabilities Monitoring Network, 11 Sites, United States, 2012. MMWR Surveill Summ 2016;65(No. SS-3):1–23.
7  Autism and Developmental Disabilities Monitoring Network Surveillance Year 2006 Principal Investigators. Prevalence of autism spectrum disorders – Autism and Developmental Disabilities Monitoring Network, United States, 2006. MMWR Surveill Summ 2009;58(No. SS-10):1–20.
8  Autism and Developmental Disabilities Monitoring Network Surveillance Year 2010 Principal Investigators. Prevalence of autism spectrum disorder among children aged eight years—Autism and Developmental Disabilities Monitoring Network, 11 sites, United States, 2010. MMWR Surveill Summ 2014;63(No. SS-2).
9  Blei DM, Ng AY, Jordan MI. Latent dirichlet allocation. Journal of Machine Learning Research. 2003;3(Jan):993-1022.
10 Ramage D, Hall D, Nallapati R et al. Labeled LDA: A supervised topic model for credit attribution in multi-labeled corpora. In Proceedings of the 2009 Conference on Empirical Methods in Natural Language Processing: Volume 1-Volume 1 2009 Aug 6 (pp. 248-256). Association for Computational Linguistics.
11 Dumais ST, Furnas GW, Landauer TK, et al. Using latent semantic analysis to improve access to textual information. In Proceedings of the SIGCHI conference on Human factors in computing systems 1988 May 1 (pp. 281-285). ACM.
12 Rennie JD, Shih L, Teevan J, et al. Tackling the poor assumptions of naive bayes text classifiers. In Proceedings of the 20th International Conference on Machine Learning (ICML-03) 2003 (pp.616-623).
13 Cortes C, Vapnik V. Support-vector networks. Machine Learning. 1995 Sep 1;20(3):273-97.
14 Wang S, Manning CD. Baselines and bigrams: Simple, good sentiment and topic classification. In Proceedings of the 50th Annual Meeting of the Association for Computational Linguistics: Short Papers-Volume 2 2012 Jul 8 (pp. 90-94). Association for Computational Linguistics.

15. Mesnil G, Mikolov T, Ranzato MA, et al. Ensemble of generative and discriminative techniques for sentiment analysis of movie reviews. arXiv preprint arXiv:1412.5335. 2014 Dec 17.
16. Joulin A, Grave E, Bojanowski P, et al. Bag of tricks for efficient text classification. arXiv preprint arXiv:1607.01759. 2016 Jul 6.
17. Benjamini Y, Yekutieli D. The control of the false discovery rate in multiple testing under dependency. Annals of statistics. 2001 Aug 1:1165-88.
18. Pedregosa F, Varoquaux G, Gramfort A, et al. Scikit-learn: Machine learning in Python. Journal of Machine Learning Research. 2011;12(Oct):2825-30.
19. Walt SV, Colbert SC, Varoquaux G. The NumPy array: a structure for efficient numerical computation. Computing in Science & Engineering. 2011 Mar;13(2):22-30.
20. Abadi M, Agarwal A, Barham P, et al. Tensorflow: Large-scale machine learning on heterogeneous distributed systems. arXiv preprint arXiv:1603.04467. 2016 Mar 14.
21. R Core Team. R: A language and environment for statistical computing.
22. Platt J. Probabilistic outputs for support vector machines and comparisons to regularized likelihood methods. Advances in large margin classifiers. 1999 Mar 26;10(3):61-74.
23. Pascanu R, Mikolov T, Bengio Y. On the difficulty of training recurrent neural networks. In International Conference on Machine Learning 2013 Feb 13 (pp. 1310-1318).
24. Hochreiter S, Schmidhuber J. Long short-term memory. Neural computation. 1997 Nov 15;9(8):1735-80.
25. Cho K, Van Merriënboer B, Bahdanau D, et al. On the properties of neural machine translation: Encoder-decoder approaches. arXiv preprint arXiv:1409.1259. 2014 Sep 3.
26. Kim Y. Convolutional neural networks for sentence classification. arXiv preprint arXiv:1408.5882. 2014 Aug 25.
27. Zhang X, Zhao J, LeCun Y. Character-level convolutional networks for text classification. In Advances in Neural Information Processing Systems 2015 (pp. 649-657).
28. Denil M, Demiraj A, Kalchbrenner N, Blunsom P, et al. Modelling, visualising and summarising documents with a single convolutional neural network. arXiv:1406.3830. 2014 Jun 15.
29. Kalchbrenner N, Blunsom P. Recurrent convolutional neural networks for discourse compositionality. arXiv preprint arXiv:1306.3584. 2013 Jun 15.
30. Tang D, Qin B, Liu T. Document Modeling with Gated Recurrent Neural Network for Sentiment Classification. In EMNLP 2015 Sep 17 (pp. 1422-1432).
31. Dai AM, Olah C, Le QV. Document embedding with paragraph vectors. arXiv preprint arXiv:1507.07998. 2015 Jul 29.
32. Newcombe RG. Improved confidence intervals for the difference between binomial proportions based on paired data. Statistics in medicine. 1998 Nov 30;17(22):2635-50.
33. Leisenring W, Alono T, Pepe MS. Comparisons of predictive values of binary medical diagnostic tests for paired designs. Biometrics. 2000 Jun 1;56(2):345-51.
34. Demšar J. Statistical comparisons of classifiers over multiple data sets. Journal of Machine Learning Research. 2006;7(Jan):1-30.

# SUPPLEMENT

*S0. Overview of Hyperparameter Optimization Procedures*

After splitting the data randomly into a training set and a validation set (note: the seed for this split was not used as a seed for any of the 10 train-test splits reported in the primary experiment), we use 3 kinds of optimization procedures to tune the hyperparameters for our models: grid search (LSA and LDA); recursive feature elimination (RFE; random forest); and a Bayesian method for Gaussian process optimization (GPO; all other models). We implement grid search using NumPy; RFE using a combination of NumPy and scikit-learn; and GPO using the GPyOpt library for Python. All GPO used the default parameters for the methods.bayesian_optimization module in GPyOpt, but we changed the number of iterations for which the process was allowed to run for certain models. We report these changes below, along with full descriptions of tuning procedures and selected hyperparameters for each model below.

*S1. Latent Dirichlet Allocation*
For the decomposition, we use the default hyperparameters for LDA in scikit-learn. In combination with the linear SVM, then, there are 2 hyperparameters for our combined LDA+SVM classifier: the number of topics $n_{topics}$ for the LDA; and the C parameter for the SVM. We consider $n_{topics}$ in {5, 10, 15, 20, 30} and discrete values for C in {0.001, 0.01, 0.1, 1, 2, 8, 16}.. After grid search to minimize classification error on the validation set, the best value of $n_{topics}$ was 30, and the best value for C was 8.

*S2. Latent Semantic Analysis*
For the decomposition, we use the TruncatedSVD class in scikit-learn. In combination with the linear SVM, then, there are 2 hyperparameters for our combined LSA+SVM classifier: the dimensionality $d$ of the singular value decomposition for the document-term matrix; and the C parameter for the SVM. We consider $d$ in {10, 25, 50, 100, 200} and discrete values of C in {0.001, 0.01, 0.1, 1, 2, 8, 16}. After grid search to minimize classification error on the validation set, the best value for $d$ was 100, and the best value for C was 0.001.

*S3. Multinomial Naïve Bayes*
For our multinomial naïve Bayes classifier, we use the MultinomialNB class from scikit-learn, and we searched for continuous values of the smoothing parameter alpha in [0.0001, 1.0]. After GPO ($n_{iter}$=50) to minimize classification error on the validation set, the best value of alpha was 0.032683.

*S4. Linear SVM*
For our SVM, with use the LinearSVC class from scikit-learn, and we searched for discrete values of C in {0.0001, 0.001, 0.01, 0.1, 1, 2, 5}. Minimizing classification error on the validation set, both grid search and GPO ($n_{iter}$=50) settled on 0.0001 as the best value. We also note here that applying TF-IDF weights to the bigram feature vectors did not improve classification accuracy.

*S5. Random forest*
For our random forest, we use the RandomForestClassifier class from scikit-learn with n_estimators set to 1,000 and the rest of the hyperparameters to their defaults. As in Maenner et al. 2016, we chose a different threshold for classification than 0.50. To select our threshold, we examined classification accuracy for cutoffs in (0.01, 0.99) in steps of 0.01 on the initial validation set, and chose 0.47, which produced the highest accuracy. We used this threshold for the rest of our experiments.

As in Maenner et al. 2016, we found feature selection to improve model performance, and we explored 2 procedures: recursive feature elimination (RFE); and non-recursive feature elimination (nRFE). For both procedures, we begin by fitting the model to the training data, and we create a new document-term matrix containing only the top 250 most important features from this initial fit (this is mainly for the sake of convenience; stepping the model down from the full 860,493 bigrams would take an enormous amount of time). Then, we use the elimination procedures to find the number of remaining features $n_{top}$ that produces the highest accuracy on the validation set. For RFE, this means stepping down from the trimmed set of 250 features in increments of 10, stopping once the specified number of features has been reached; and for nRFE, this means simply specifying the number of features to keep and discarding all of the others. We allowed both methods to search for $n_{top}$ between 10 and 200 in steps of 10 (so possible values were 10, 20, 30, etc.). After this process, RFE chose a value for $n_{top}$ of 120, and nRFE a value of 130. To evaluate these two procedures, we tested each across the 10 train-test splits in the main experiment. The performance was similar, but nRFE posted the higher mean accuracy (87.10 vs. 86.83), and so its results are what we report.

For all the random forest models, we used TF-IDF-weighted count-valued bigram feature vectors as inputs.

*S6. NB-SVM*
We formulate our NB-SVM in the same way as Wang and Manning (2012), i.e. by converting binarized (where any count > 1 is converted to 1) bigram features to NB features and then allowing for interpolation between these and the SVM. For the SVM, we use the LinearSVC class from scikit-learn, and we modify it with custom code in NumPy. Like Wang and Manning, we keep squared hinge loss and the L2 penalty for the SVM, and we use a value of 1.0 for the NB smoothing parameter α. Thus, there are only two hyperparameters to tune here: the C parameter for the SVM, and the interpolation parameter β. We search for continuous values of β in (0.0, 1.0) and discrete values of C in {0.001, 0.01, 1.0, 2, 4} (we note that Wang and Manning use β=0.25 and C=1.0). After GPO ($n_{iter}$=30) to minimize classification error on the validation set, we settled on β of 1.0 and C of 0.001, which is equivalent to an SVM with NB features and no interpolation

*S7. Neural networks*
We formulate our neural networks as deep averaging classifiers where the first layer is a lookup matrix for the feature embeddings, and the second is a fully-connected layer with a sigmoid activation. In our first model $NN_{sum}$, the word embeddings are added together before being passed to the second layer, and in our second model $NN_{avg}$, they are averaged (the latter approach is used by the fastText classifier). In both models, we also apply dropout to the output of the embedding layer to prevent overfitting. For training, the networks were optimized using Adam (Kingma 2014) to minimize binary cross-entropy loss on the training data until loss on the test data had not increased for some a fixed number of epochs. This number, which we call the patience parameter, was treated as a hyperparameter and tuned along with several other hyperparameters using the procedures described below.

As our hyperparameters, we considered dropout probability $p$, patience, embedding size $e$, minibatch size, and learning rate. For $NN_{avg}$, we searched for continuous $p$ in (0.0, 0.9), discrete patience in {5, 10, 15, 20, 25}, discrete embedding size in {64, 128, 256, 512}, discrete minibatch size in {32, 64, 128, 256}, and discrete learning rate in (0.0001, 0.001, 0.01, 0.1); for $NN_{sum}$ the ranges for the same parameters were (0.0, 0.9), {2, 5, 10}, {64, 128, 256, 512}, {32, 64, 128, 256}, and {0.00001, 0.0001, 0.001}, respectively (this model was more likely to overfit, so we lowered the learning rate and decreased patience accordingly). After GPO($n_{iter}$=20), the optimal hyperparameters for NN$_{avg}$ were {0.75, 10, 64, 32, 0.001}, and the optimal hyperparameters for NN$_{sum}$ were {0.86, 5, 64, 256, 0.00001}.

Both models were trained on unigrams only and not bigrams; including the latter did not improve performance.

**REFERENCES**

Kingma DP, Ba J. Adam: A method for stochastic optimization. arXiv preprint arXiv:1412.6980. 2014 Dec 22.